\pdfoutput=1

\documentclass[11pt]{article}

\usepackage[preprint]{acl}

\usepackage{times}
\usepackage{latexsym}

\usepackage{comment}

\usepackage[T1]{fontenc}

\usepackage[utf8]{inputenc}

\usepackage{microtype}

\usepackage{inconsolata}

\usepackage{graphicx}

\usepackage{xspace}
\usepackage{tcolorbox}
\usepackage{subfig}
\usepackage{paralist}
\usepackage{enumitem}

\usepackage{xcolor}
\usepackage{soul}
\usepackage[normalem]{ulem}

\setlist[itemize]{itemsep=0.02cm,topsep=0.2cm}
\setlist[enumerate]{itemsep=0.02cm,topsep=0.2cm}

\definecolor{incorrect}{HTML}{FFBCBC}
\definecolor{notcheckable}{HTML}{E9D2FF}
\definecolor{misleading}{HTML}{FFF79F}
\definecolor{other}{HTML}{DDDDDD}

\newcommand{\factgenie}{\texttt{factgenie}\xspace}
\tcbset{on line, 
        boxsep=4pt, left=0pt,right=0pt,top=0pt,bottom=0pt,
        colframe=white}

\title{\factgenie: A Framework for Span-based Evaluation of Generated Texts}

\author{Zdeněk Kasner \quad Ondřej Plátek \quad  {\bf Patrícia Schmidtová}  \\ {\bf Simone Balloccu} \quad {\bf Ondřej Dušek} \\
        Institute of Formal and Applied Linguistics \\ 
        Faculty of Mathematics and Physics, Charles University \\
        \texttt{\$\{surname\}@ufal.mff.cuni.cz}}

\begin{document}
\maketitle
\global\csname @topnum\endcsname 0
\global\csname @botnum\endcsname 0
\begin{abstract}
  We present
  \factgenie: a framework for annotating and visualizing word spans in textual model outputs. Annotations can capture various span-based phenomena such as semantic inaccuracies or irrelevant text. With \factgenie, the annotations can be collected both from human crowdworkers and large language models. Our framework consists of a web interface for data visualization and gathering text annotations, powered by an easily extensible codebase.\footnote{Code is available at \url{https://github.com/kasnerz/factgenie/}. System demonstration video: \url{https://youtu.be/CsVcCGvOzPY}.}
\end{abstract}

%


\section{Introduction}
The fluency of texts generated by large language models (LLMs) is reaching the level of human-written texts. However, the texts generated by LLMs still contain various types of errors such as incorrect claims, claims not grounded in the input, or irrelevant statements. For precise and fine-grained evaluation of model outputs, it is necessary to identify these errors on the level of word spans. There are two major ways to collect the span annotations: using either human \citep{thomsonGoldStandardMethodology2020} or LLM-based annotators \cite{kocmi-federmann-2023-gemba,kasnerReferenceBasedMetricsAnalyzing2024}.



\begin{figure}[t]
  \subfloat[Custom visualization of the input data.]{%
    \includegraphics[clip,width=\columnwidth]{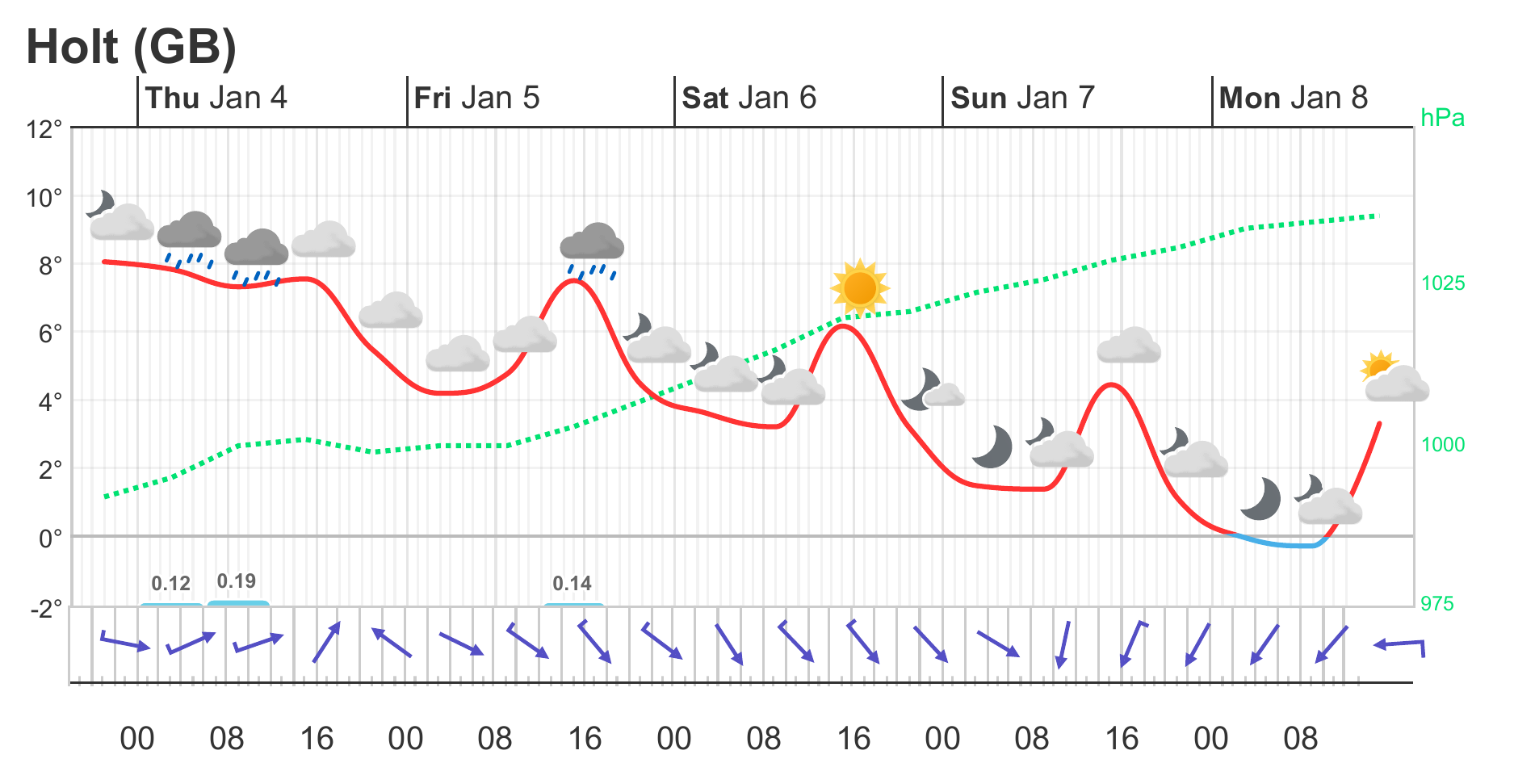}%
    \label{fig:vis}
  }
  \footnotesize{\newline}
  \subfloat[Annotated model output.]{%
    \includegraphics[clip,width=\columnwidth]{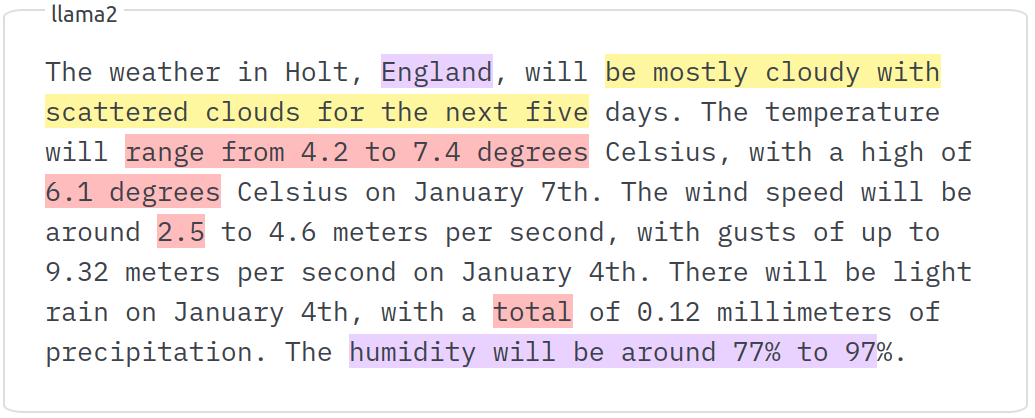}%
    \label{fig:ann}
  }
  \caption{Elements from the \factgenie user interface: (a) custom visualization of the input data, (b) the corresponding LLM output with span annotations. The highlight colors correspond to custom annotation categories defined for the annotation process (\tcbox[colback=incorrect,tcbox raise=-1mm]{\null} = incorrect fact, \tcbox[colback=notcheckable,tcbox raise=-1mm]{\null} = fact not checkable, \tcbox[colback=misleading,tcbox raise=-1mm]{\null} = misleading fact).}
  \label{fig:llama2_annotated}
\end{figure}

None of the existing NLP error annotation platforms are suitable for gathering and visualizing word-level annotations from both human and LLM-based annotators.
Some platforms are limited to specific tasks like machine translation \citep{klejch2015mt} and retrieval-augmented generation \citep{es-etal-2024-ragas}.
Other platforms are more flexible but allow either only human \citep{federmann-2018-appraise,doccano} or only LLM \citep{dalvi-etal-2024-llmebench} annotations.
Systems supporting both annotation modalities typically include humans as post-editors only \citep{kim-etal-2024-meganno} and existing evaluation or visualization platforms require externally pre-annotated data \citep{trebuna-dusek-2023-visuallm,masson-etal-2024-textbi,fittschen-etal-2024-annoplot}.

\begin{figure*}[!htp]
  \centering
  \includegraphics[trim=4cm 4cm 4cm 4cm,clip,width=\textwidth]{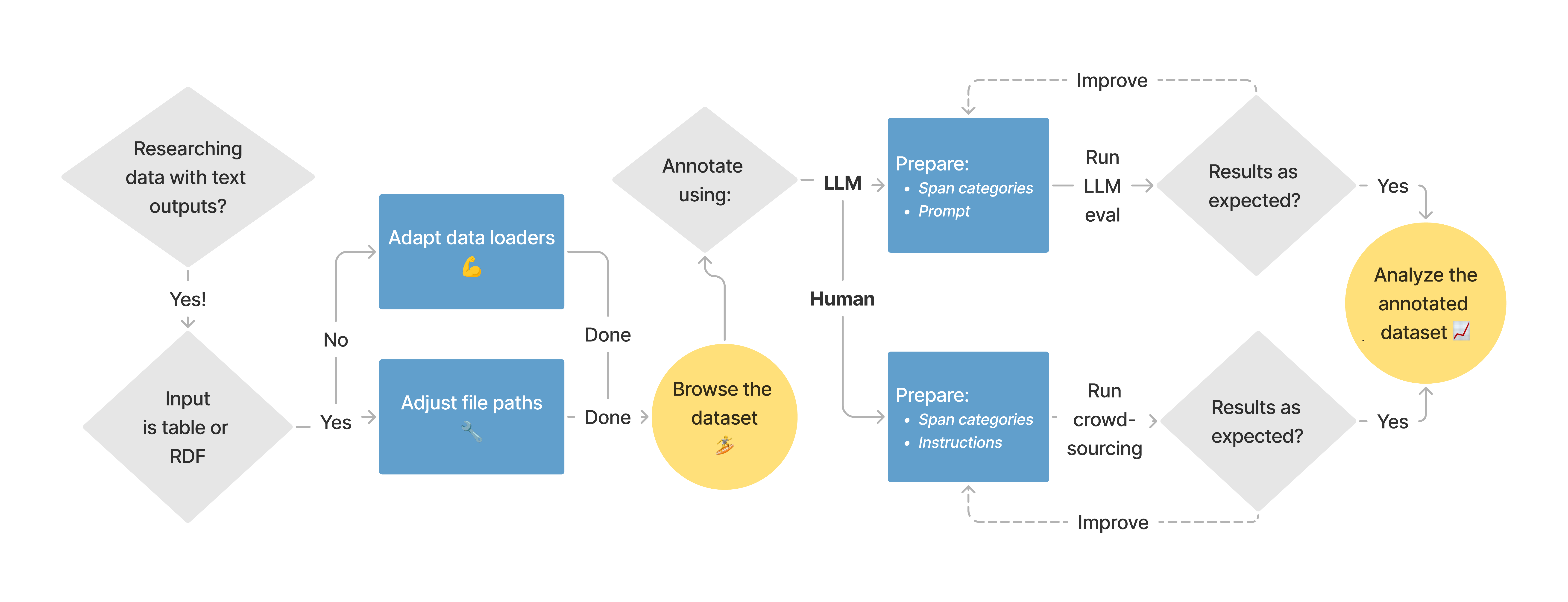}
  \caption{\factgenie workflow. Actions needed for using \factgenie for custom tasks are shown in blue rectangles.}
  \label{fig:workflow}
\end{figure*}

The lack of suitable tools for span-based error annotation
motivated us to develop \factgenie, a lightweight and customizable framework that enables collecting annotations from both humans and LLMs.
Specifically, \factgenie can be used both to (a) collect annotations from human workers through crowdsourcing services and (b) collect annotations by prompting an LLM through an API. Besides that, \factgenie can be used for visualizing the input data and the corresponding model outputs.

The software design of \factgenie targets researchers, who can easily self-host and customize it for individual experiments. The benefits of \factgenie include:
\begin{itemize}
  \item Visualization of input data and model outputs with a few lines of code,
  \item Ready-made web interface for collecting annotations from crowdsourcing services,
  \item Support for gathering model-based annotations from multiple LLM APIs,
  \item Tools for managing and visualizing collected annotations.


\end{itemize}
\label{sec:intro}

\section{Framework}
\label{sec:framework}

Software-wise, \factgenie is a combination of a \href{https://flask.palletsprojects.com}{Flask} backend and an HTML-based frontend. The frontend is powered by \href{https://getbootstrap.com/docs/5.3/}{Boostrap 5.3} and \href{https://jquery.com}{jQuery}, additionally using the \href{https://github.com/SuLab/YPet}{YPet} library for collecting span annotations. For visualizing the example input data, we use \href{https://github.com/niklasf/python-tinyhtml/tree/master}{TinyHTML} and \href{https://www.highcharts.com}{Highcharts.JS}.\footnote{In principle, \factgenie can render datasets using any custom HTML code and JS libraries.}

\autoref{fig:llama2_annotated} shows an example with weather data and the corresponding model-generated weather forecast. The model output was annotated for errors through \factgenie. Note that the colors and labels of text span annotation categories can be customized for each set of annotations.

The framework can be used as-is or customized to cover a wide range of tasks and needs with minimal effort. To load and preview a new dataset, researchers first need to write a data loader class. Existing data loaders include various visualizations of tabular, RDF, and JSON data. As shown in \autoref{fig:workflow}, loading a dataset in a supported format can be as easy as changing a path to the data on the file system. To add a custom dataset type, the researcher must extend the \texttt{Dataset} class. Once the dataset is loaded, \factgenie allows data inspection and rapid prototyping of LLM annotations and crowdsourcing campaigns.

\section{Human Annotations}
\label{sec:human}
To collect error annotation from human crowdworkers, researchers typically build custom web interfaces. With \factgenie, researchers can easily build an annotation interface in four steps:
\begin{enumerate}
  \item Define the campaign parameters (annotation span categories, number of examples per annotator, etc.),
  \item Write instructions for the annotators,
  \item Host \factgenie on a public URL,
  \item Redirect the annotators to the running \factgenie instance.
\end{enumerate}
The interface can be previewed for internal testing throughout the process.
As shown in \autoref{fig:workflow}, \factgenie provides the feedback necessary for debugging and improving the evaluation campaign by an immediate visualization of the collected annotations.

\section{LLM Annotations}
\label{sec:llm}
It is useful to obtain annotations from LLMs in the same format as from human annotators. For that, \factgenie provides a lightweight wrapper for model APIs.\footnote{We currently support the \href{https://github.com/ollama/ollama/blob/main/docs/api.md}{Ollama API} for self-hosted LLMs and the \href{https://openai.com/api/}{OpenAI API} for cloud LLMs.} The process of collecting annotations from LLMs consists of the following steps:
\begin{enumerate}
  \item Define the campaign parameters (annotation span categories, model decoding parameters, API endpoint, etc.)
  \item Write the prompt and system message for the model,\footnote{The prompt needs to instruct the model to produce JSON with a specific structure. Note that the APIs we support can ensure decoding JSON output, see, e.g., \url{https://platform.openai.com/docs/guides/json-mode}.}
  \item Run the LLM annotation inference.
\end{enumerate}

Similarly to human annotations (\autoref{sec:human}), the evaluation progress can be monitored and immediately visualized.

\section{Roadmap}
The development of \factgenie is ongoing and open to external developers. We are currently working on facilitating the management of evaluation campaigns by adding an option to set-up the evaluation campaign from the web interface in addition to configuration files. In the future, we plan to add more ready-made classes for data loaders, model APIs, and crowdsourcing services.


\section*{Acknowledgements}

This work was funded by the European Union (ERC, NG-NLG, 101039303) and Charles University projects GAUK 40222 and SVV 260~698. It used resources of the LINDAT/CLARIAH-CZ Research Infrastructure (Czech Ministry of Education, Youth, and Sports project No. LM2018101).

\bibliography{custom}

\begin{thebibliography}{12}
\providecommand{\natexlab}[1]{#1}

\bibitem[{Dalvi et~al.(2024)Dalvi, Hasanain, Boughorbel, Mousi, Abdaljalil, Nazar, Abdelali, Chowdhury, Mubarak, and Ali}]{dalvi-etal-2024-llmebench}
Fahim Dalvi, Maram Hasanain, Sabri Boughorbel, Basel Mousi, Samir Abdaljalil, Nizi Nazar, Ahmed Abdelali, Shammur~Absar Chowdhury, Hamdy Mubarak, and Ahmed Ali. 2024.
\newblock \href {https://aclanthology.org/2024.eacl-demo.23} {{{LLM}e{B}ench}: A flexible framework for accelerating {{LLM}s} benchmarking}.
\newblock In \emph{Proceedings of the 18th Conference of the European Chapter of the Association for Computational Linguistics: System Demonstrations}, pages 214--222, St. Julians, Malta.

\bibitem[{ES et~al.(2024)ES, James, Anke, and Schockaert}]{es-etal-2024-ragas}
Shahul ES, Jithin James, Luis~Espinosa Anke, and Steven Schockaert. 2024.
\newblock \href {https://aclanthology.org/2024.eacl-demo.16} {{RAGAs}: Automated evaluation of retrieval augmented generation}.
\newblock In \emph{Proceedings of the 18th Conference of the European Chapter of the Association for Computational Linguistics, {EACL} 2024 - System Demonstrations, St. Julians}, pages 150--158, Malta.

\bibitem[{Federmann(2018)}]{federmann-2018-appraise}
Christian Federmann. 2018.
\newblock \href {https://aclanthology.org/C18-2019/} {Appraise evaluation framework for machine translation}.
\newblock In \emph{{COLING} 2018, The 27th International Conference on Computational Linguistics: System Demonstrations}, pages 86--88, Santa Fe, New Mexico.

\bibitem[{Fittschen et~al.(2024)Fittschen, Fischer, Br{\"u}hl, Spahr, Lysa, and Le}]{fittschen-etal-2024-annoplot}
Elisabeth Fittschen, Tim Fischer, Daniel Br{\"u}hl, Julia Spahr, Yuliia Lysa, and Phuoc~Thang Le. 2024.
\newblock \href {https://aclanthology.org/2024.eacl-demo.12} {{{A}nno{P}lot}: Interactive visualizations of text annotations}.
\newblock In \emph{Proceedings of the 18th Conference of the European Chapter of the Association for Computational Linguistics: System Demonstrations}, pages 106--114, St. Julians, Malta.

\bibitem[{Kasner and Du{\v s}ek(2024)}]{kasnerReferenceBasedMetricsAnalyzing2024}
Zden{\v e}k Kasner and Ond{\v r}ej Du{\v s}ek. 2024.
\newblock \href {http://arxiv.org/abs/2401.10186} {Beyond {{{Traditional} Benchmarks}}: {{{Analyzing} Behaviors}} of {{{Open} {LLMs}}} on {{{Data-to-Text} Generation}}}.
\newblock In \emph{Proceedings of the 62nd Annual Meeting of the Association for Computational Linguistics (Volume 1: Long Papers)}.
\newblock To appear.

\bibitem[{Kim et~al.(2024)Kim, Mitra, Li~Chen, Rahman, and Zhang}]{kim-etal-2024-meganno}
Hannah Kim, Kushan Mitra, Rafael Li~Chen, Sajjadur Rahman, and Dan Zhang. 2024.
\newblock \href {https://aclanthology.org/2024.eacl-demo.18} {{{MEGA}nno+}: A {Human-{LLM}} collaborative annotation system}.
\newblock In \emph{Proceedings of the 18th Conference of the European Chapter of the Association for Computational Linguistics: System Demonstrations}, pages 168--176, St. Julians, Malta.

\bibitem[{Klejch et~al.(2015)Klejch, Avramidis, Burchardt, and Popel}]{klejch2015mt}
Ondrej Klejch, Eleftherios Avramidis, Aljoscha Burchardt, and Martin Popel. 2015.
\newblock \href {http://ufal.mff.cuni.cz/pbml/104/art-klejch-et-al.pdf} {{MT-ComparEval}: Graphical evaluation interface for machine translation development}.
\newblock \emph{Prague Bull. Math. Linguistics}, 104:63--74.

\bibitem[{Kocmi and Federmann(2023)}]{kocmi-federmann-2023-gemba}
Tom Kocmi and Christian Federmann. 2023.
\newblock \href {https://doi.org/10.18653/V1/2023.WMT-1.64} {{GEMBA-MQM:} detecting translation quality error spans with {GPT-4}}.
\newblock In \emph{Proceedings of the Eighth Conference on Machine Translation, {WMT} 2023}, pages 768--775, Singapore.

\bibitem[{Masson et~al.(2024)Masson, Sallaberry, Bessagnet, Le~Parc~Lacayrelle, Roose, and Agerri}]{masson-etal-2024-textbi}
Maxime Masson, Christian Sallaberry, Marie-Noelle Bessagnet, Annig Le~Parc~Lacayrelle, Philippe Roose, and Rodrigo Agerri. 2024.
\newblock \href {https://aclanthology.org/2024.eacl-demo.1} {{{T}ext{BI}}: An interactive dashboard for visualizing multidimensional {NLP} annotations in social media data}.
\newblock In \emph{Proceedings of the 18th Conference of the European Chapter of the Association for Computational Linguistics: System Demonstrations}, pages 1--9, St. Julians, Malta.

\bibitem[{Nakayama et~al.(2018)Nakayama, Kubo, Kamura, Taniguchi, and Liang}]{doccano}
Hiroki Nakayama, Takahiro Kubo, Junya Kamura, Yasufumi Taniguchi, and Xu~Liang. 2018.
\newblock \href {https://github.com/doccano/doccano} {{doccano}: Text annotation tool for human}.
\newblock Software available from \url{https://github.com/doccano/doccano}.

\bibitem[{Thomson and Reiter(2020)}]{thomsonGoldStandardMethodology2020}
Craig Thomson and Ehud Reiter. 2020.
\newblock \href {https://doi.org/10.18653/V1/2020.INLG-1.22} {A gold standard methodology for evaluating accuracy in data-to-text systems}.
\newblock In \emph{Proceedings of the 13th International Conference on Natural Language Generation, {INLG} 2020}, pages 158--168, Dublin, Ireland.

\bibitem[{Trebu{\v{n}}a and Dušek(2023)}]{trebuna-dusek-2023-visuallm}
Franti{\v{s}}ek Trebu{\v{n}}a and Ondřej Dušek. 2023.
\newblock \href {https://aclanthology.org/2023.inlg-demos.3} {{{V}isua{LLM}}: Easy web-based visualization for neural language generation}.
\newblock In \emph{Proceedings of the 16th International Natural Language Generation Conference: System Demonstrations}, pages 6--8, Prague, Czechia.

\end{thebibliography}




\end{document}